\documentclass{bmvc2k}

\title{Resolving Semantic Confusions for Improved Zero-Shot Detection}

\addauthor{Sandipan Sarma}{sandipan.sarma@iitg.ac.in}{1}
\addauthor{Sushil Kumar}{sushil13129340@gmail.com}{1}
\addauthor{Arijit Sur}{arijit@iitg.ac.in}{1}

\addinstitution{
 Indian Institute of Technology Guwahati \\
 Guwahati, India
}

\runninghead{S. Sarma, S. Kumar, A. Sur}{Resolving SC for ZSD}

\def\eg{\emph{e.g}\bmvaOneDot}

\usepackage{amsfonts}
\usepackage{wrapfig}
\usepackage{multirow, multicol}
\usepackage{adjustbox}
\usepackage{subfigure}
\usepackage{hyperref}
\hypersetup{
	colorlinks=true,
	linkcolor=blue,
	filecolor=magenta,      
	urlcolor=magenta,
}
\begin{document}

\maketitle

\begin{abstract}
Zero-shot detection (ZSD) is a challenging task where we aim to recognize and localize objects simultaneously, even when our model has not been trained with visual samples of a few target (``unseen'') classes. Recently, methods employing generative models like GANs have shown some of the best results, where unseen-class samples are generated based on their semantics by a GAN trained on seen-class data, enabling vanilla object detectors to recognize unseen objects. However, the problem of semantic confusion still remains, where the model is sometimes unable to distinguish between semantically-similar classes. In this work, we propose to train a generative model incorporating a triplet loss that acknowledges the degree of dissimilarity between classes and reflects them in the generated samples. Moreover, a cyclic-consistency loss is also enforced to ensure that generated visual samples of a class highly correspond to their own semantics. Extensive experiments on two benchmark ZSD datasets -- MSCOCO and PASCAL-VOC -- demonstrate significant gains over the current ZSD methods, reducing semantic confusion and improving detection for the unseen classes. Codes and models will be released at \url{https://github.com/sandipan211/ZSD-SC-Resolver}.
\end{abstract}

\section{Introduction}
\label{sec:intro}

The idea of {\em zero-shot detection} (ZSD)~\cite{Demirel2018ZeroShotOD, rahman2018zero, bansal2018zero, zhu2019zero} has been recently introduced with the aim of transferring knowledge about some ``seen'' classes to the ``unseen'' with the help of semantic information relating these classes. At test time, trained ZSD models are evaluated in two settings -- (i) test images contain unseen objects only (conventional ZSD), and (ii) test images can have objects from both seen and unseen classes ({\em generalized ZSD or GZSD}). While many of the ZSD approaches are inspired by the success of zero-shot recognition methods focusing on improving visual-semantic alignment~\cite{bansal2018zero, rahman2018zero, Demirel2018ZeroShotOD, rahman2018polarity, mao2020zero}, others tried leveraging additional information in the form of textual descriptions~\cite{li2019zero} and synthesizing unseen samples using generative methods~\cite{zhao2020gtnet, zhu2020don, hayat2020synthesizing}, achieving state-of-the-art results. Interestingly, most of these approaches suffer from the problem of {\em semantic confusion}, where the knowledge transfer between the seen and unseen classes bridged by semantic representations is not discriminative enough at times to distinguish between semantically-similar classes. This results in low average precisions for the ZSD models, which could prove to be catastrophic when deployed in real-world environments in the future, \eg in medical imaging systems where a high false-positive rate can negatively impact their reliability, or in underwater explorations where low luminance and turbidity can misguide the detection of marine debris.     

We propose a generative method for ZSD, which aims to resolve semantic confusion by utilizing a triplet loss~\cite{schroff2015facenet} while training the feature synthesizer. Specifically, we use Faster-RCNN~\cite{ren2016faster} as a backbone object detector that can be trained on images containing only seen class objects. Fixed-size feature vectors for these objects are used to train a conditional Wasserstein GAN~\cite{arjovsky2017wasserstein} (cWGAN) regularized by a classification loss, following the success of~\cite{xian2018feature} in the zero-shot recognition task. In order to ensure diversity among the synthesized features, we use a regularization term~\cite{mao2019mode} that alleviates the issue of mode collapse in conditional GANs. However, such a cWGAN only learns how to synthesize image features conditioned upon class semantics and does not account for the degree of dissimilarity between object classes while learning to synthesize their features. Hence we introduce a triplet loss that can primarily help in learning discriminative features for the semantically-similar classes, resolving semantic confusion whenever these synthesized features are utilized for our detection pipeline ahead. 
Moreover, we explicitly aim to maintain the consistency between the synthesized visual features and semantics of the corresponding class by incorporating a cyclic-consistency loss enforcing the synthesized visual features to reconstruct their semantics. The trained cWGAN is used to generate unseen class features, which are used to update the classifier of the pretrained Faster-RCNN, empowering it to detect unseen-class objects as well. Since the performance of this classifier is directly related to the quality of synthesized features used as inputs for training, accounting for inter-class dissimilarity and visual-semantic consistency can impact the performance of the detector. Moreover, using a generative method also minimizes the {\em hubness problem}~\cite{radovanovic2010hubs, dinu2014improving, shigeto2015ridge, gupta2020multi, zhao2020gtnet, hayat2020synthesizing}.

We summarize our contributions in this work as follows: (i) we propose using a triplet loss with a flexible semantic margin (refer to Sec.~\ref{sec:sem-confusion}) while training a feature synthesizer (cWGAN) which generates unseen object features conditioned upon class semantics and enables our backbone object detector to detect both seen and unseen objects, resolving semantic confusion between similar objects; (ii) visual-semantic consistency is maintained during feature generation, ensuring generated features correspond well to their semantic counterparts; (iii) extensive experiments are performed on two benchmark ZSD datasets (MSCOCO and PASCAL-VOC) which show that our method remains comparable to the best existing methods in case of PASCAL-VOC, but comprehensively beats these methods on the more challenging and bigger dataset MSCOCO in both conventional ZSD and GZSD settings.

\section{Related Work}
\label{sec:related}
Motivated by zero-shot classification (ZSC) methods~\cite{rohrbach2010helps, rohrbach2011evaluating, kankuekul2012online, akata2015label, frome2013devise, akata2015evaluation, romera2015embarrassingly, kodirov2017semantic, xian2016latent, zhang2015zero, changpinyo2016synthesized, norouzi2013zero, zhu2019learning, vyas2020leveraging, schonfeld2019generalized, narayan2020latent, mishra2018generative, xian2018feature, xian2019f, feng2020transfer, tang2021zero, liu2020novel, Thong2020BiasGZSL, huang2020multi}, the more challenging task of ZSD~\cite{rahman2018zero, bansal2018zero, zhu2019zero, Demirel2018ZeroShotOD} started gaining attention since 2018. The initial works seek to improve visual-semantic alignment and extend popular ZSC frameworks like ConSE~\cite{norouzi2013zero} with trusted object detectors like Faster-RCNN. In these works, usually projection functions are learned~\cite{rahman2018zero, Demirel2018ZeroShotOD, mao2020zero} for capturing seen-unseen and visual-semantic relationships. However, recent methods have shown that such projection-based strategies can be improved. Contrastive losses designed with respect to semantic vectors and put into action within a joint intermediate embedding space for the visual features and semantic vectors in~\cite{yan2022semantics}, and a polarity loss~\cite{rahman2018polarity} which refines the noisy semantic vectors and explicitly maximizes the gap between positive and negative predictions, are two examples of such methods. 

Some methods additionally target the problem of background-unseen confusion (BUC), where ZSD models confuse unseen objects with background at test time due to low objectness scores for unseen objects. Additional data from external sources is used for obtaining a vocabulary with classes belonging to neither seen nor unseen classes in~\cite{bansal2018zero}, encoding an idea about the background classes. Vocabulary atoms~\cite{chua2009nus} enrich the semantic space with a diverse set of linguistic concepts and help relate to the visual features better in~\cite{rahman2018polarity}. The same vocabulary is used in~\cite{zheng2020background} along with a background-learnable RPN for detecting objects.

Rather than focusing only on visual-semantic alignment, several methods explore some other limitations in ZSD using different data structures and multi-modal approaches. Unseen-class {\em localization} gets priority in~\cite{zhu2019learning}, where predicting class attributes is a side-task and the produced bounding boxes utilize both visual and semantic information. In a multi-modal approach,~\cite{li2019zero} uses unit-level and word-level attention from a language branch for weighing the outputs of a visual branch for detecting objects. A GCN-based~\cite{welling2016semi} approach is taken up in~\cite{yan2020semantics}, utilizing a graph construction module and two semantics-preserving graph propagation modules. Transformer-based encoder-decoder networks have been used in~\cite{zheng2021zero}, achieving a stronger ability to deal with BUC and recall unseen objects.

With generative networks~\cite{goodfellow2014generative, DBLP:journals/corr/KingmaW13} minimizing the hubness problem~\cite{radovanovic2010hubs}, a few ZSD methods have also turned to such networks. A conditional VAE is employed in~\cite{zhu2020don} for synthesizing unseen features used to update the confidence predictor of a YOLO~\cite{redmon2017yolo9000} detector, pre-trained with seen objects. In another work~\cite{zhao2020gtnet}, three separate GANs are used to generate visual features with both intra-class variance and IoU variance. Instead,~\cite{hayat2020synthesizing} encourages a unified GAN model that generates discriminative object features and ensures feature diversity. Inspired by the potential of generative methods, we also employ one for our ZSD model (Fig.~\ref{fig:zsd-model}). However, our work differs majorly from the previous generative approaches in the use of visual-semantic cyclic consistency loss and in addressing the problem of semantic confusion, which none of the previous works do. We compare our results in Sec.~\ref{sec:expt} with methods following all these aforementioned approaches. 

\section{Method}
\label{sec:method}
{\bf Problem Setting:} We formally define ZSD here. Let $\mathcal{C}^s = \{ 1,2,...S \}$ and $\mathcal{C}^u = \{ S+1, S+2,...S+U \}$ be label sets of $S$ number of seen and $U$ number of unseen classes respectively, such that $\mathcal{C}^s \cap \mathcal{C}^u = \phi$. Moreover, in object detection, a concept of {\em background class} must be identified too -- so total class labels become $S+U+1$.  Let the training data $\mathcal{D}^{tr} = \{I_m, \{O_m^i \}_{i = 1}^{N_m}  \}_{m = 1}^{M}$ consist of $M$ images, having $N_m$ objects with class annotations in the set $\{O_m^i \}_{i = 1}^{N_m}$ for an image $I_m$. The $i^{th}$ object in $I_m$ is annotated as $O_m^i = \{B_m^i, c_m^i\}$, where $B_m^i = \{x_m^i, y_m^i, w_m^i, h_m^i \}$ denotes the bounding-box coordinates and $c_m^i \in \mathcal{C}^s$. 
Moreover, the semantic descriptions for the seen and unseen classes ($d$-dimensional word embeddings~\cite{mikolov2013distributed, pennington2014glove, joulin-etal-2017-bag}) are given as $\mathcal{P}^{s} \in \mathbb{R}^{S \times d}$ and $\mathcal{P}^{u} \in \mathbb{R}^{U \times d}$ respectively. At test time, images containing objects from both $\mathcal{C}^s$ and $\mathcal{C}^u$ can be given, and the goal would be to predict bounding boxes for every foreground object, along with their class labels. 

\noindent {\bf Backbone object detector:} 
We use a Faster-RCNN~\cite{ren2016faster} $\Phi_{\tt frcn}$ with a ResNet-101~\cite{he2016deep} pretrained on ImageNet~\cite{russakovsky2015imagenet} classes (except the {\em overlapping unseen classes}~\cite{xian2018zero}) as a feature extractor for the input images, yielding a convolutional feature map ({\em ConvMap}). A {\em Region Proposal Network} (RPN) predicts objectness scores for different {\em region proposals} in the ConvMap. For each of the top $N_p$ proposals adjudged as {\em foreground} by the RPN, fixed-size feature vectors are extracted from its projection on the ConvMap via RoI pooling. These are fed to fully-connected layers that branch into a classification module $\phi_{\tt frcn-c}$ which classifies a proposal into one of $N+1$ classes ($N$ foreground classes and a ``background'' class), and a bounding-box regression module $\phi_{\tt frcn-r}$ which regresses over box coordinates for object localization. $\Phi_{\tt frcn}$ is trained on $\mathcal{D}^{tr}$, and used to extract object-level visual features $f^i_m \in \mathcal{F}_m $ for every seen object in an image $I_m$.

\subsection{The feature synthesizer}
\label{sec:synthesizer}
\begin{figure}[t]
	\centering
	\includegraphics[width=0.7\textwidth]{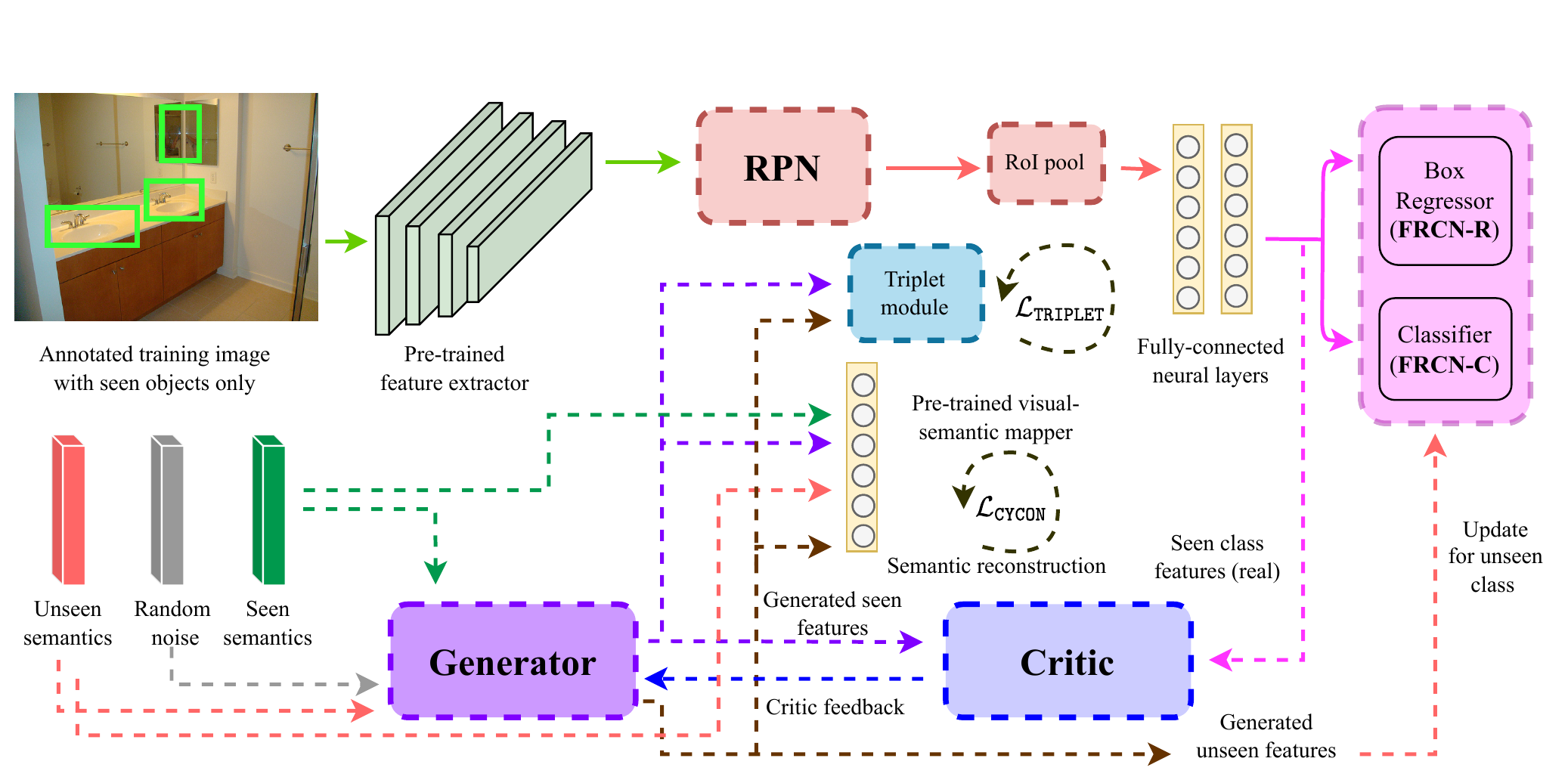}
	\caption{{\bf Model architecture for the proposed ZSD model.} The {\em solid} arrows show the workflow in the backbone object detector part, whereas the {\em dashed arrows} show the workflow regarding the feature synthesizer (cWGAN), optimizing the objective function in Eq.~\ref{eq:total_loss}.}  \
	\label{fig:zsd-model}
	
\end{figure}

The baseline architecture of the proposed model is depicted in Fig.~\ref{fig:zsd-model} (similar GAN also used in a prior ZSC task~\cite{xian2018feature}). The collection of object features from all training images $\mathcal{F}^s = \{\mathcal{F}_m \}_{m = 1}^M$ along with the corresponding seen-class labels serve as {\em real data} for a conditional WGAN~\cite{arjovsky2017wasserstein} (cWGAN). A generator network learns a mapping function $\mathcal{G}: \mathcal{Z} \times \mathcal{P}^s \longrightarrow \mathcal{F}^s$ that takes $p \in \mathcal{P}^s$ and $z \in \mathcal{Z}$ as inputs and learns the underlying distributions of the visual features from $\mathcal{F}^s$, relates them to the corresponding semantics. Here, $z \sim \mathcal{N}(0, 1) \in \mathbb{R}^{z_d}$ is a random noise vector sampled from a Gaussian distribution. During training, the generator $\mathcal{G}$ generates class-wise seen object features ({\em fake data}), feeds them to a critic network $\mathcal{Q}$ for scoring its {\em realness} or {\em fakeness} and recalibrates itself based on the feedback from $\mathcal{Q}$ to make the generated feature distribution as close to real distribution as possible, minimizing the loss:

\begin{align}
	\label{eq:wgan}
	\mathcal{L}_{\tt WGAN} = \mathbb{E}[\mathcal{Q}(f, c)] - \mathbb{E}[\mathcal{Q}(\Tilde{f}, c)] + \lambda \mathbb{E}[(|| \nabla_{\hat{f}} \mathcal{Q}(\hat{f}, c) ||_2 -1)^2]
\end{align}

where the first two terms represent the critic loss in WGAN and the third term represents a gradient penalty~\cite{gulrajani2017improved}, with $\lambda$ being the penalty coefficient. $\Tilde{f} = \mathcal{G}(z, p)$ denotes the generated feature, and $\hat{f} = \rho f + (1 - \rho)\Tilde{f}$, with $\rho \sim U(0,1)$~\cite{xian2018feature, gulrajani2017improved}. We add a regularization term~\cite{xian2018feature} that enforces discriminative feature generation using: 
\begin{align}
	\label{eq:cls}
	\mathcal{L}_{\tt CLS} = -\mathbb{E}[\log \mathbf{p}(c|\mathcal{G}(z, p); \phi_{cls}^{sm})]
\end{align}

where $\mathbf{p}(y|\mathcal{G}(.))$ denotes a classification probability given by a linear softmax classifier $\phi_{cls}^{sm}$ pretrained on $\mathcal{F}^s$. Additionally, inspired by~\cite{hayat2020synthesizing}, we consider the impact of individual noise vectors on feature generation and enhance feature diversity to prevent the problem of mode collapse~\cite{salimans2016improved} by including a mode-seeking regularization term~\cite{mao2019mode}:

\begin{align}
	\label{eq:ms}
	\mathcal{L}_{\tt MS} = \mathbb{E}[ || \mathcal{G}(z_1, p) - \mathcal{G}(z_2, p) ||_1 / || z_1 - z_2 ||_1]
\end{align}

\subsection{Triplet loss for resolving semantic confusion}
\label{sec:sem-confusion}

Semantic confusion is a major hindrance in ZSL tasks, where semantically-similar classes are hard to distinguish at test time, dropping prediction scores for the seen and unseen classes. In the ZSC task, this problem has been addressed using triplet loss-based methods~\cite{akata2015label, frome2013devise}. In this work, we leverage a modified triplet loss for our ZSD task. However, our execution of the triplet loss is different from the ones used in ZSC, where multi-modal triplets are formed with visual features as ``anchors'' and class semantics as ``positive'' or ``negative'' matches, and compatibility scores indicate multi-modal similarity. On the contrary, our triplets are strictly visual feature-based, having the form <$\Tilde{f}_a, \Tilde{f}_p, \Tilde{f}_n$>, where $\Tilde{f}_a$ is the anchor feature of a cWGAN-generated sample from a given class, $\Tilde{f}_p$ is a positive match of the same class as $\Tilde{f}_a$, and $\Tilde{f}_n$ is a negative match of a different class. Instead of compatibility scores, we focus on similarity in the visual space spanned by our triplets and optimize:

\begin{align}
	\label{eq:triplet}
	\mathcal{L}_{\tt TRIPLET} = \max \{ 0, (d(\Tilde{f}_a, \Tilde{f}_p) - d(\Tilde{f}_a, \Tilde{f}_n) + \Delta) \}
\end{align}

where $d(.,.)$ is the Euclidean distance. Our motivation is that the quality of the features generated on the basis of class semantics heavily impacts the classification ability of $\phi_{\tt frcn-c}$ at a later stage -- therefore, accounting for inter-class dissimilarities explicitly in the visual space can address semantic confusion during the feature generation phase itself.
In addition, Eq.~\ref{eq:triplet} uses a flexible semantic margin $\Delta$ acquired as a pre-computed value from a rescaled Mahalanobis distance matrix with respect to class semantics~\cite{cacheux2019modeling}, instead of keeping it at a fixed, constant value~\cite{akata2015label, frome2013devise}. This is because we want the generator to acknowledge the degree of dissimilarity between two classes and reflect them in the generated features by accounting for the first and second-order statistics of the class semantics. 

To the best of our knowledge, such an approach has not been considered in ZSD yet. Moreover, we recognize the fact that utilizing triplet loss for a generative ZSD approach like ours could prove quite beneficial as our cWGAN can explicitly learn to reduce confusion between semantically-similar classes, producing generated features that are robust to such semantic similarities.

\begin{table}[t]
	\caption{ZSD and GZSD performance of various methods on MSCOCO in terms of mAP and Recall@100 (RE@100) at an IoU threshold of 0.5. HM denotes the harmonic mean of seen and unseen results for GZSD. The best results and second-best results are shown in \textit{red} and \textit{blue} respectively. Results are shown for the 65/15 split of MSCOCO.}
	\label{tab:zsd1-coco-mapre}
	
	\centering
	\begin{tabular}{|c|c|c|c|c|c|}
		\hline
		{\bf Metric} & {\bf Method} & {\bf ZSD} & \multicolumn{3}{|c|}{{\bf GZSD}} \\ 
		\cline{4-6}
		&  &  & {\bf Seen} & {\bf Unseen} & {\bf HM} \\ \hline
		\multirow{7}{*}{mAP}
		& PL~\cite{rahman2018polarity}  & 12.40& 34.07 & 12.40 & 18.18 \\
		
		& BLC~\cite{zheng2020background} &14.70 &36.00 &13.10 &19.20 \\
		& ACS-ZSD~\cite{mao2020zero} &15.34 &- &- &- \\
		& SUZOD\cite{hayat2020synthesizing} &17.30  &  37.40 & \textcolor{blue}{17.30} & \textcolor{blue}{23.65} \\
		
		& ZSDTR~\cite{zheng2021zero} & 13.20 & {\bf \textcolor{red}{40.55}} & 13.22 & 20.16 \\
		& ContrastZSD~\cite{yan2022semantics}  & \textcolor{blue}{18.60} & \textcolor{blue}{40.20} & 16.50 & 23.40 \\
		
		\cline{2-6}
		
		& {\bf Ours}  & {\bf \textcolor{red}{20.10}} & 37.40 & {\bf \textcolor{red}{20.10}} & {\bf \textcolor{red}{26.15}} \\ 
		\hline
		
		\multirow{7}{*}{RE@100} 
		& PL~\cite{rahman2018polarity} & 37.72& 36.38 & 37.16 & 36.76 \\
		
		& BLC~\cite{zheng2020background} & 54.68 & 56.39 & 51.65 & 53.92 \\
		& ACS-ZSD\cite{mao2020zero} &47.83 &- &- &- \\
		& SUZOD\cite{hayat2020synthesizing}  & \textcolor{blue}{61.40} &  58.60 & \textcolor{blue}{60.80} & 59.67 \\
		& ZSDTR~\cite{zheng2021zero} & 60.30 &  {\bf \textcolor{red}{69.12}} &  59.45 & \textcolor{blue}{ 61.12} \\
		& ContrastZSD~\cite{yan2022semantics}  &59.50& \textcolor{blue}{62.90} & 58.60 & 60.70 \\
		\cline{2-6}
		
		& {\bf Ours} & {\bf \textcolor{red}{65.10}} & 58.60 & {\bf \textcolor{red}{64.00}} & {\bf \textcolor{red}{61.18}} \\ \hline  
	\end{tabular}
\end{table}

\subsection{Cyclic consistency between the visual and semantic spaces}
\label{sec:cyc-con}
The cWGAN assumes that generated object features for a class conditioned upon its semantics would have a distribution close to the real features of that class. This stems from an implicit assumption that the visual and semantic distributions for that class are relatively similar. However, this can bias the generated features of the unseen classes towards semantically-similar seen classes on which the cWGAN is trained, negatively affecting $\phi_{\tt frcn-c}$ at the next stage. 
Hence, we implement a cyclic consistency module forcing the generated features to reconstruct the class semantics based on which they were generated in the first place. A visual-semantic mapper $\mathcal{M}: \mathcal{F}^s \longrightarrow \mathcal{P}^s$ is first trained on seen data, which is an MLP that learns to map object features of seen classes to their semantic counterparts, minimizing a semantic reconstruction loss.
During cWGAN training, the synthesized features are passed to the pretrained $\mathcal{M}$ which reconstructs the semantics from these features, and a reconstruction loss is optimized as:

\begin{align}
	\label{eq:cycon}
	\mathcal{L}_{\tt CYCON} =  \underset{p \sim \mathcal{P}^s}{\mathbb{E}}[|| p - \mathcal{M}(\mathcal{G}(z, p)) ||_2^2] 
	+ 
	\underset{p \sim \mathcal{P}^u}{\mathbb{E}}[|| p - \mathcal{M}(\mathcal{G}(z, p)) ||_2^2]
\end{align}

The overall objective function becomes:
\begin{align}
	\label{eq:total_loss}
	\underset{\mathcal{G}}{min} \; \underset{\mathcal{Q}}{max} \, \alpha_1 \mathcal{L}_{\tt WGAN} + \alpha_2 \mathcal{L}_{\tt CLS} + \alpha_3 \mathcal{L}_{\tt MS} + \alpha_4 \mathcal{L}_{\tt CYCON} + \alpha_5 \mathcal{L}_{\tt TRIPLET},
\end{align}
making the generated features diverse and robust to semantic confusions and visual-semantic inconsistencies. Once the cWGAN is trained, we train a classifier $\phi_{cls}^{u}$ using the generated unseen features. The learned weights are provided to $\phi_{\tt frcn-c}$ to make it capable of classifying unseen visual features.

\begin{table}[t]
	\caption{ Class-wise average precisions (APs) on unseen classes (ZSD) from MSCOCO with an IoU threshold of 0.5. The best results and second-best results are shown in \textit{red} and \textit{blue}.}
	
	\centering
	\label{tab:zsd1-class_wise_coco}
	\begin{adjustbox}{max width=\textwidth}
		\begin{tabular}{|c *{17}{|p{4ex}}|}
			\hline
			{\bf Method} &
			\rotatebox[origin=c]{90}{{\bf Overall}} &
			\rotatebox[origin=c]{90}{aeroplane} &
			\rotatebox[origin=c]{90}{train} &
			\rotatebox[origin=c]{90}{p.meter} &
			\rotatebox[origin=c]{90}{cat} &
			\rotatebox[origin=c]{90}{bear} &
			\rotatebox[origin=c]{90}{suitcase} &
			\rotatebox[origin=c]{90}{frisbee} &
			\rotatebox[origin=c]{90}{snowbrd.} &
			\rotatebox[origin=c]{90}{fork} &
			\rotatebox[origin=c]{90}{sandwich} &
			\rotatebox[origin=c]{90}{hot dog} &
			\rotatebox[origin=c]{90}{toilet} &
			\rotatebox[origin=c]{90}{mouse} &
			\rotatebox[origin=c]{90}{toaster} &
			\rotatebox[origin=c]{90}{hair drier} 
			\\ \hline

			PL\cite{rahman2018polarity} &12.40&\textcolor{blue}{20.0}&\textcolor{blue}{48.2}&0.63&28.3&13.8&\textcolor{blue}{12.4}&{\bf \textcolor{red}{21.8}}&15.1&8.9&8.5&\textcolor{blue}{0.87}&5.7&0.04&{\bf \textcolor{red}{1.7}}&\textcolor{blue}{0.03}\\ 
			
			ACS-ZSD\cite{mao2020zero} &15.34&8.72&25.5&{\bf \textcolor{red}{6.59}}&40.8&\textcolor{blue}{54.0}&9.55&\textcolor{blue}{10.6}&26.8&\textcolor{blue}{16.4}&11.0&\textcolor{red}{4.99}&7.83&{\bf \textcolor{red}{6.21}}&\textcolor{blue}{1.32}&0.0\\
			
			SUZOD\cite{hayat2020synthesizing} &\textcolor{blue}{17.30}&17.8&46.3&\textcolor{blue}{0.7}&\textcolor{blue}{63.1}&41.0&10.5&
			0.7&\textcolor{blue}{30.2}&{\bf \textcolor{red}{16.5}}&\textcolor{blue}{17.6}&0.0&\textcolor{blue}{13.4}&{1.6}&0.4&{\bf \textcolor{red}{0.2}} \\ \hline
			
			{\bf Ours}   &{\bf \textcolor{red}{20.10}}&{\bf \textcolor{red}{22.9}}&{\bf \textcolor{red}{53.3}}&0.6&{\bf \textcolor{red}{64.9}}&{\bf \textcolor{red}{54.3}}&{\bf \textcolor{red}{13.2}}&1.2&{\bf \textcolor{red}{31.2}}&15.7&{\bf \textcolor{red}{22.6}}&0.0&{\bf \textcolor{red}{17.5}}&\textcolor{blue}{2.7}&0.7&{\bf \textcolor{red}{0.2}} \\ \hline
			
		\end{tabular}
	\end{adjustbox}
	
\end{table}

\begin{table}[t]
	\caption{mAP (in \%) for PASCAL-VOC dataset. The unseen classes are shown in \underline{italic}. The best and second-best results are shown in {\em red} and {\em blue}.}
	\centering
	\label{tab:zsd1-voc_class_table}
	\begin{adjustbox}{max width=\textwidth}
		\begin{tabular}{|c *{23}{|p{4ex}}|}
			\hline
			
			{\bf Method} &
			\rotatebox[origin=c]{90}{{\bf seen}} &
			\rotatebox[origin=c]{90}{{\bf unseen}} &
			\rotatebox[origin=c]{90}{aeroplane} &
			\rotatebox[origin=c]{90}{bicycle} &
			\rotatebox[origin=c]{90}{bird} &
			\rotatebox[origin=c]{90}{boat} &
			\rotatebox[origin=c]{90}{bottle} &
			\rotatebox[origin=c]{90}{bus} &
			\rotatebox[origin=c]{90}{cat} &
			\rotatebox[origin=c]{90}{chair} &
			\rotatebox[origin=c]{90}{cow} &
			\rotatebox[origin=c]{90}{d.table} &
			\rotatebox[origin=c]{90}{horse} &
			\rotatebox[origin=c]{90}{motorbike} &
			\rotatebox[origin=c]{90}{person} &
			\rotatebox[origin=c]{90}{p.plant} &
			\rotatebox[origin=c]{90}{sheep} &
			\rotatebox[origin=c]{90}{tvmonitor} &
			\rotatebox[origin=c]{90}{\textit{{\underline{car}}}} &
			\rotatebox[origin=c]{90}{\textit{{\underline{dog}}}}&
			\rotatebox[origin=c]{90}{\textit{{\underline{sofa}}}} &
			\rotatebox[origin=c]{90}{\textit{{\underline{train}}}}\\ \hline
			
			HRE~\cite{Demirel2018ZeroShotOD} &65.6&54.2&70.0&73.0&76.0&54.0&42.0&{\bf \textcolor{red}{86.0}}&64.0&40.0&54.0&{\bf \textcolor{red}{75.0}}&80.0&80.0&75.0&34.0&69.0&{\bf \textcolor{red}{79.0}}&55.0&82.0&55.0&26.0\\ 
			
			PL~\cite{rahman2018polarity} &63.5&62.1&74.4&71.2&67.0&50.1&50.8&67.6&84.7&44.8&68.6&39.6&74.9&76.0&\textcolor{blue}{79.5}&39.6&61.6&66.1&{\bf \textcolor{red}{63.7}}&87.2&53.2&44.1\\
			
			SAN~\cite{rahman2018zero} &69.6&57.6&71.4&78.5&74.9&61.4&48.2&\textcolor{blue}{76.0}&89.1&51.1&{\bf \textcolor{red}{78.4}}&61.6&\textcolor{blue}{84.2}&76.8&76.9&42.5&71.0&71.7&\textcolor{blue}{56.2}&85.3&{\bf \textcolor{red}{62.6}}&26.4\\ 
			BLC~\cite{zheng2020background} &{\bf \textcolor{red}{75.1}} &55.2&78.5&83.2&\textcolor{blue}{77.6}&{\bf \textcolor{red}{67.7}}&{\bf \textcolor{red}{70.1}}&75.6&87.4&\textcolor{blue}{55.9}&\textcolor{blue}{77.5}&\textcolor{blue}{71.2}&{\bf \textcolor{red}{85.2}}&82.8&77.6&{\bf \textcolor{red}{56.1}}&{\bf \textcolor{red}{77.1}}&\textcolor{blue}{78.5}&43.7&86.0&\textcolor{blue}{60.8}&30.1\\ 
			
			SUZOD~\cite{hayat2020synthesizing} & \textcolor{blue}{74.7}& {\bf \textcolor{red}{63.1}}&{\bf \textcolor{red}{80.6}}&{\bf \textcolor{red}{84.2}}&{\bf \textcolor{red}{78.7}}&\textcolor{blue}{66.7}&\textcolor{blue}{67.6}&74.0&{\bf \textcolor{red}{91.5}}&{\bf \textcolor{red}{61.1}}&75.0&63.4&82.6&{\bf \textcolor{red}{85.7}}&{\bf \textcolor{red}{84.9}}&\textcolor{blue}{47.8}&\textcolor{blue}{76.9}&74.8&55.2&\textcolor{blue}{92.3}&59.0&{\bf \textcolor{red}{46.1}} \\ \hline
			{\bf Ours}  & \textcolor{blue}{74.7}& \textcolor{blue}{62.7}&\textcolor{blue}{80.4}&\textcolor{blue}{84.1}&{\bf \textcolor{red}{78.7}}&66.6&\textcolor{blue}{67.6}&73.8&\textcolor{blue}{91.4}&{\bf \textcolor{red}{61.1}}&75.0&63.7&83.1&\textcolor{blue}{85.5}&{\bf \textcolor{red}{84.9}}&47.3&76.7&74.7&55.6&{\bf \textcolor{red}{92.6}}&57.5&\textcolor{blue}{45.0} \\ \hline
			
		\end{tabular}
	\end{adjustbox}
\end{table}

\section{Experiments}
\label{sec:expt}
\subsection{Datasets and evaluation metrics} \label{sec:datasets-evalMet}
We evaluate our proposed method extensively on two commonly used datasets in ZSD -- MSCOCO and PASCAL-VOC. MSCOCO~\cite{lin2014microsoft} is a large-scale dataset containing annotated images from 80 object classes, for which we use the seen/unseen split of 65/15 provided by~\cite{rahman2018polarity}. PASCAL-VOC 2007/2012~\cite{everingham2010pascal} contains annotated images from 20 object classes, for which we use the 16/4 split provided in~\cite{Demirel2018ZeroShotOD}. We follow~\cite{hayat2020synthesizing} while obtaining the sets of training and testing images. Following~\cite{bansal2018zero} and~\cite{rahman2018polarity}, we use Recall@100 (RE@100) and mean average precision (mAP) as evaluation metrics and report our results at an IoU of 0.5. In the GZSD setting, we follow~\cite{rahman2018polarity} and show the mAP on seen and unseen classes and consider their harmonic mean (HM) as the overall performance metric.

\begin{figure}[t]
	\centering
	\includegraphics[width=\textwidth]{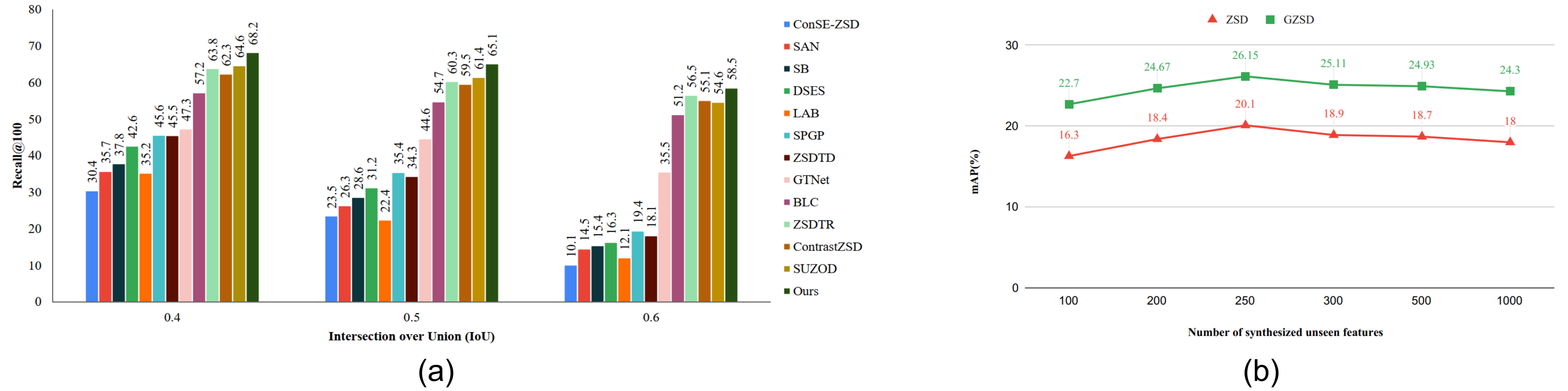}
	\caption{(a) Comparison of recall@100 for various IoU thresholds on MSCOCO; (b) mAP variation depending on the number of synthesized unseen features in case of MSCOCO (unseen mAP considered for ZSD and HM of seen and unseen mAP considered for GZSD).}  
	\label{fig:recall-ious-n-synthesized}
\end{figure}

\subsection{Implementation details}
\label{sec:zsd1-imple}
For the RPN, anchor bounding boxes with an IoU $\geq$ 0.7 are regarded as \textit{foreground} , whereas those with IoU $\leq$ 0.3 are regarded as \textit{background}, yielding $N_p = 2000$ proposals for each image with an NMS threshold of 0.7. The backbone Faster-RCNN model is trained on seen data first for 12 epochs and 4 epochs on the GPU for MSCOCO and PASCAL respectively. Classifiers and bounding-box regressors are both fully-connected neural layers. The cWGAN takes 300-dimensional FastText embedding vectors~\cite{joulin-etal-2017-bag} as class semantics and 1024-dimensional object features extracted using RoI pooling layer of the Faster-RCNN (pretrained on seen data) as {\em real features}. The generator and critic networks are implemented as single-layered neural networks with 4096 hidden units. For Eq.~\ref{eq:total_loss}, we empirically set the hyperparameter values as $\alpha_1 = 1.0, \alpha_2 = 0.01, \alpha_3 = 0.01, \alpha_4 = 0.01$ and $\alpha_5 = 0.1$. 
The cWGAN is trained for 55 epochs, and the weights from the best epoch are used further. Adam optimizer is used for both $\mathcal{G}$ and $\mathcal{Q}$ networks, with a learning rate of 0.0005 and mini-batch size of 128. Generated features for both seen and unseen classes are checked for their consistency via Eq.~\ref{eq:cycon}, and also used for constructing all possible triplets {\em online}~\cite{schroff2015facenet} for Eq.~\ref{eq:triplet}.

\subsection{Results: Comparison with the State-of-the-art}
\label{sec:results}

\begin{figure}[t]
	\centering
	\includegraphics[width=0.8\textwidth]{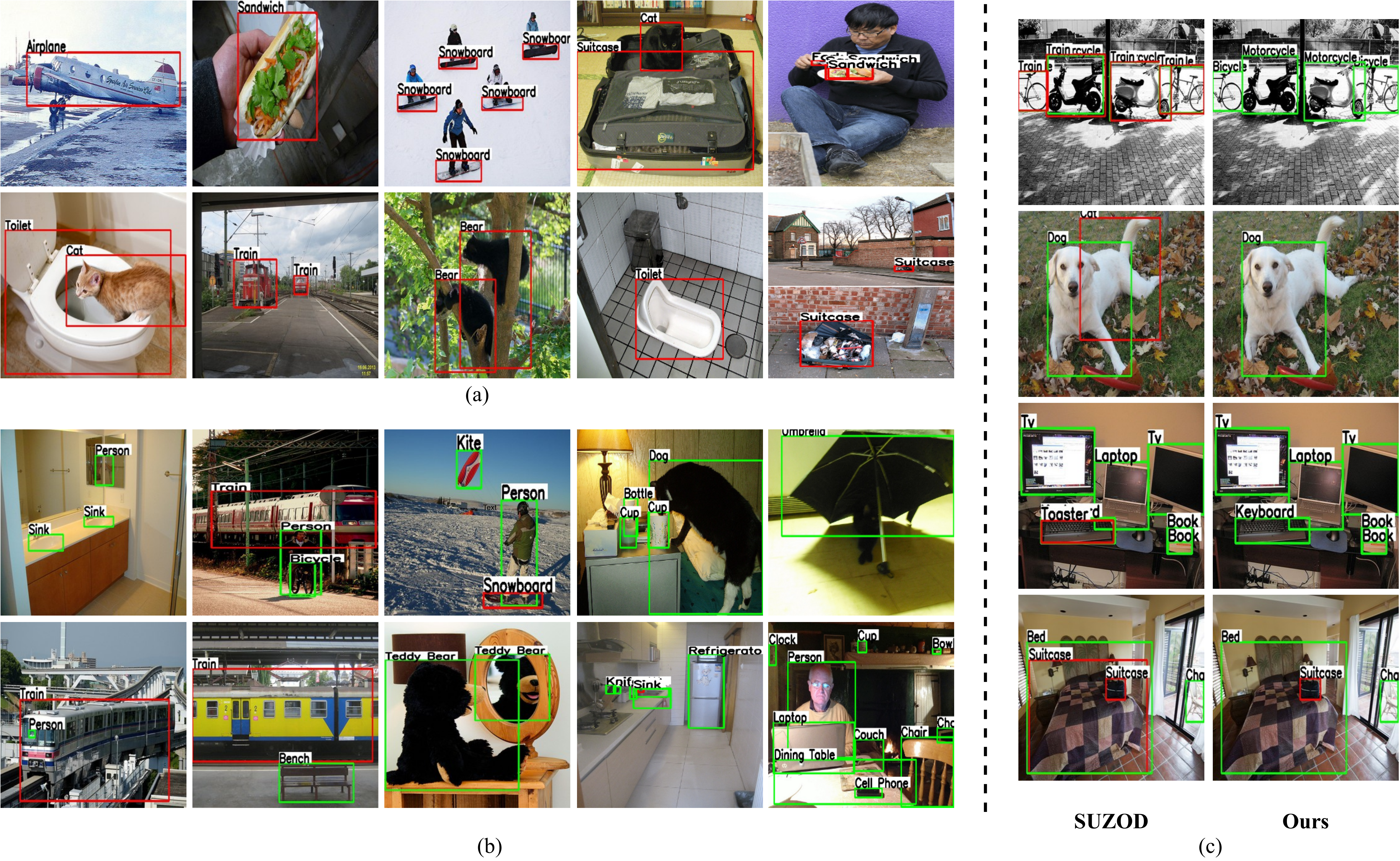}
	\caption{Qualitative results on MSCOCO (best viewed in zoom). (a) ZSD results; (b) GZSD results; (c) Comparison of our GZSD results with the current state-of-the-art (SUZOD~\cite{hayat2020synthesizing}). The seen and unseen class objects are detected within {\em green} and {\em red} boxes.}
	\label{fig:mscoco-results}
\end{figure}

\begin{figure}[t]
	\centering
	\includegraphics[width=0.9\textwidth]{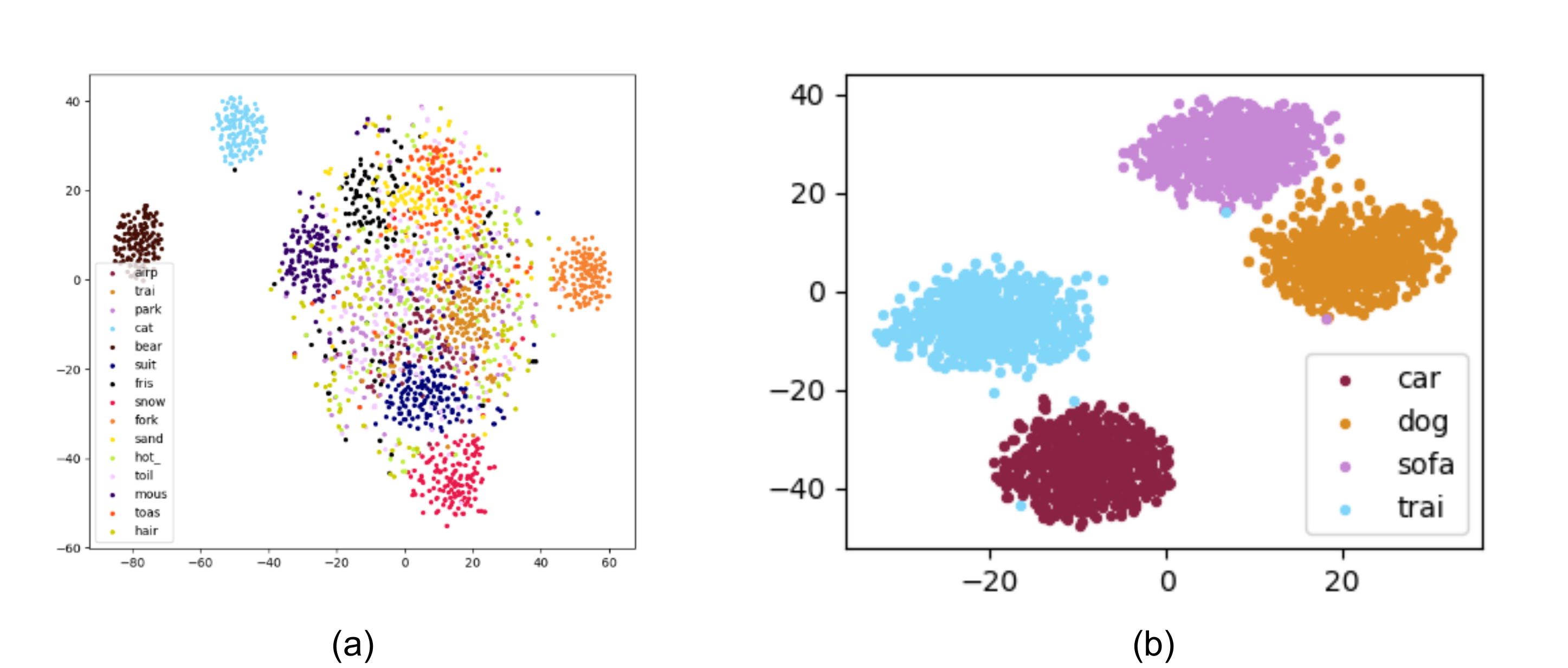}
	\caption{t-SNE visualization~\cite{van2014accelerating} of the generated features for unseen classes in (a) MSCOCO and (b) PASCAL-VOC.}
	\label{fig:t-sne}
\end{figure}

\noindent {\bf Results on MSCOCO.} (i) \underline{{\em ZSD setting}}: Table~\ref{tab:zsd1-coco-mapre} shows that our method achieves a {\bf relative mAP gain of 8\%} over the next-best method~\cite{yan2022semantics}, and a {\bf relative RE@100 gain of 6\%} over the next-best method~\cite{hayat2020synthesizing}. This reflects the superior performance of generative methods over others, especially because we get rid of the hubness problem while augmenting visual data for the unseen classes in a cycle-consistent manner.
\begin{wraptable}{r}{0.6\textwidth}
	\caption{Impact of including different loss terms for training the feature synthesizer on mAP (in \%) during evaluation on the MSCOCO dataset.}
	\label{tab:zsd1-ablation_study}
	\centering
	\begin{tabular}{ccccc|c} 
		\hline
		$\mathcal{L}_{\tt WGAN}$ & $\mathcal{L}_{\tt MS}$ & $\mathcal{L}_{\tt CLS}$ & $\mathcal{L}_{\tt CYCON}$ & $\mathcal{L}_{\tt TRIPLET}$ & mAP \\ [0.5ex] 
		\hline\hline\
		\checkmark & \checkmark & \checkmark &\checkmark  &  & 18.4 \\ 
		\hline
		
		\checkmark &  & \checkmark   & \checkmark & \checkmark& 18.5 \\
		\hline
		
		\checkmark & \checkmark &  & \checkmark & \checkmark& 19.3 \\
		\hline
		
		\checkmark & \checkmark & \checkmark &  &\checkmark  & 19.4 \\
		\hline
		
		\checkmark & \checkmark & \checkmark & \checkmark & \checkmark & {\bf 20.1} \\
		\hline

	\end{tabular}
	
\end{wraptable}
Moreover, the gain we achieve for a challenging metric like mAP, where false positives are penalized, suggests that our triplet loss is indeed reducing semantic confusion and helping in correct classification of unseen objects. At a class-level (Tab.~\ref{tab:zsd1-class_wise_coco}), we achieve better results for most unseen classes. However, some classes like {\em parking meter}, {\em frisbee} and {\em hot dog} show decreased performance, possibly due to the unavailability of semantically-similar seen classes, affecting conditional feature generation. The same is reflected in the t-SNE~\cite{van2014accelerating} plot in Fig.~\ref{fig:t-sne}(a), where these classes do not form very well-defined clusters as compared to other unseen classes like {\em bear}, pointing out a challenging aspect of generative approaches.

Figure~\ref{fig:recall-ious-n-synthesized}(a) shows a comparison with different methods for the RE@100 values at different IoU thresholds during ZSD evaluation. Our method outperforms all methods~\cite{norouzi2013zero, rahman2018zero, bansal2018zero, yan2020semantics, li2019zero, zhao2020gtnet, zheng2020background, zheng2021zero, yan2022semantics, hayat2020synthesizing} for three IoU thresholds, even when some of the existing methods use additional semantic information from external vocabularies~\cite{bansal2018zero, li2019zero}. 

(ii) \underline{{\em GZSD setting}}: GZSD is a challenging setting as there always exists a bias towards the seen classes for the model. Table~\ref{tab:zsd1-coco-mapre} shows a {\bf relative mAP gain of 10.5\%} over the next-best method (SUZOD~\cite{hayat2020synthesizing}) considering HM. We, therefore, comprehensively beat the current state-of-the-art results for MSCOCO in both ZSD and GZSD settings. 

(iii) \underline{{\em Qualitative analysis}}: Figure~\ref{fig:mscoco-results}(a) shows our method can detect multiple unseen instances of the same object class as well as multiple objects from different classes. Objects from different viewpoints and small sizes (Fig.~\ref{fig:mscoco-results}(b)) have also been detected well. Low localization error is encountered for GZSD cases too, with good robustness against background clutter and occlusion. Figure~\ref{fig:mscoco-results}(c) compares the localization and classification abilities of our method with the state-of-the-art GZSD method~\cite{hayat2020synthesizing}. Unlike our method,~\cite{hayat2020synthesizing} succumbs to semantic confusion and wrongly detects objects like {\em bicycle, motorcycle, keyboard}, and {\em bed}.

\begin{figure}[t]
	\centering
	\includegraphics[width=0.9\textwidth]{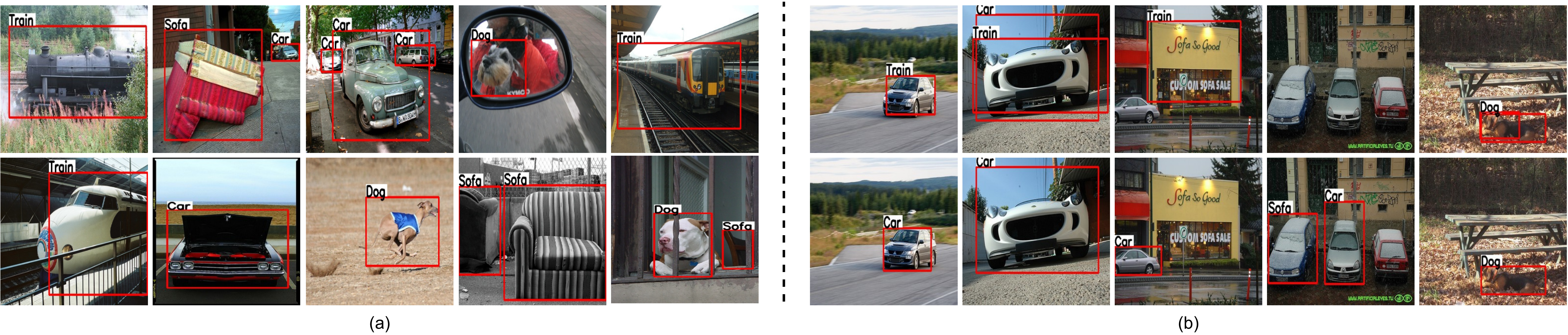}
	\caption{Qualitative results on PASCAL-VOC (best viewed in zoom). (a) Unseen detections; (b) Comparison of our results (bottom row) with SUZOD~\cite{hayat2020synthesizing} (top row).}
	\label{fig:pascal-results}
\end{figure}  

%

\noindent {\bf Results on PASCAL-VOC.} The unseen classes show improved results as compared to state-of-the-art ( SUZOD~\cite{hayat2020synthesizing}) on account of improvement in the generated unseen-class features (Fig.~\ref{fig:pascal-results}(a)). However, Tab.~\ref{tab:zsd1-voc_class_table} reports our unseen mAP is second-best to~\cite{hayat2020synthesizing} (only by 0.4\%), probably due to the relatively small size of the training data, which might not be enough to learn inter-class dissimilarities sufficiently and hence make an impact on the feature synthesizer and object detector. Nevertheless, we compare our visual results with~\cite{hayat2020synthesizing} in Fig.~\ref{fig:pascal-results}, showing that semantic confusion between {\em car} and {\em train} is common for~\cite{hayat2020synthesizing}, but not for us. The t-SNE plot of the generated unseen class features in Fig.~\ref{fig:t-sne}(b) demonstrates good separability between classes, even for semantically-similar classes like {\em car} and {\em train}.

\subsection{Ablation Studies} 
\label{sec:ablation}
\textbf{Effect of the loss components}: Table~\ref{tab:zsd1-ablation_study} shows that $\mathcal{L}_{\tt CYCON}$ provides an mAP boost as per our intuition, which indicates that generated features by the cWGAN are more discriminative and consistent with their class semantics. However, $\mathcal{L}_{\tt TRIPLET}$ has the strongest influence on ZSD, without which mAP is 18.4\% -- but jumps to 19.4\% even when $\mathcal{L}_{\tt CYCON}$ is not utilized. $\mathcal{L}_{\tt CLS}$ provides mAP boost only when included in conjunction with the other loss terms. However, optimal performance is attained when using all five loss terms for training cWGAN.

\noindent {\bf Effect of the number of generated examples}: We fix the number of generated seen features while training cWGAN and vary the number of class-wise unseen features generated by the trained cWGAN. When evaluated on MSCOCO in ZSD and GZSD settings, we find our model achieves optimal results when 250 features are generated per unseen class (Fig.~\ref{fig:recall-ious-n-synthesized}(b)). 


\section{Conclusion}
While transferring knowledge from the seen to unseen classes, most existing ZSD methods face confusion while detecting semantically similar classes. We propose a generative method that inherently eliminates the hubness problem in zero-shot conditions. Our triplet loss with a flexible semantic margin acknowledges the degree of dissimilarity between object classes while learning to synthesize discriminative object features. Moreover, a cyclic-consistency loss is enforced to maintain the visual-semantic consistencies during feature generation. Experiments and ablation studies on two challenging datasets show that we achieve state-of-the-art results, both qualitatively and quantitatively, improving upon some of the fundamental challenges the existing ZSD methods face, such as semantic confusion, high false-positive rate, and misclassification of localized objects. From a research perspective, future directions include improving the model architecture for better localization of unseen classes and reducing background-unseen confusion. From an application perspective, our ZSD model can be used as a plug-and-play module in the future for various vision applications, even in challenging environments. For instance, images captured in underwater environments can be pre-processed via image restoration techniques and fed to our ZSD model. Our model can detect novel species of fish, corals, and also help in trash detection by localizing different kinds of unseen trash objects such as plastic snack wrappers. 

\section{Acknowledgement}
This work is supported by IITG Technology Innovation and Development Foundation (IITGTI\&DF), which has been set up at IIT Guwahati as a part of the National Mission on Interdisciplinary Cyber Physical Systems (NMICPS). IITGTI\&DF is undertaking research, development and training activities on Technologies for Under Water Exploration with the financial assistance from Department of Science and Technology, India through grant number DST/NMICPS/TIH12/IITG/2020. Authors gratefully acknowledge the support provided for the present work. We also acknowledge the Department of Biotechnology, Govt. of India for the financial support for the project BT/COE/34/SP28408/2018 (for computing resources).

\bibliography{egbib}
\end{document}